%% file: root.tex
\documentclass[letterpaper, 10 pt, conference]{ieeeconf}  

\IEEEoverridecommandlockouts                              

\title{\LARGE \bf
Best of Sim and Real: Decoupled Visuomotor Manipulation via Learning Control in Simulation and Perception in Real
}

\author{Jialei Huang$^{1}$, Zhaoheng Yin$^{2}$, Yingdong Hu$^{1}$, Shuo Wang$^{1}$, Xingyu Lin$^{2}$ and Yang Gao$^{1}$
\thanks{$^{1}$Jialei Huang, Yingdong Hu, Shuo Wang and Yang Gao are with Tsinghua University, Beijing, China}%
\thanks{$^{2}$Zhaoheng Yin and Xingyu Lin are with University of California, Berkeley, CA, USA}%
}
\usepackage{graphicx}
\usepackage{amsmath}
\usepackage{amssymb}
\begin{document}

\maketitle
\thispagestyle{empty}
\pagestyle{empty}

\begin{abstract}
Sim-to-real transfer remains a fundamental challenge in robot manipulation due to the entanglement of perception and control in end-to-end learning. We present a decoupled framework that learns each component where it is most reliable: control policies are trained in simulation with privileged state to master spatial layouts and manipulation dynamics, while perception is adapted only at deployment to bridge real observations to the frozen control policy. Our key insight is that control strategies and action patterns are universal across environments and can be learned in simulation through systematic randomization, while perception is inherently domain-specific and must be learned where visual observations are authentic. Unlike existing end-to-end approaches that require extensive real-world data, our method achieves strong performance with only 10-20 real demonstrations by reducing the complex sim-to-real problem to a structured perception alignment task. We validate our approach on tabletop manipulation tasks, demonstrating superior data efficiency and out-of-distribution generalization compared to end-to-end baselines. The learned policies successfully handle object positions and scales beyond the training distribution, confirming that decoupling perception from control fundamentally improves sim-to-real transfer.
\end{abstract}

\input{intro}

\input{method}
\input{experiment}
\input{related}

\section{conclusion}

In this paper, we proposed a decoupled framework for sim-to-real transfer in manipulation, which separates perception and control learning. Our key insight that control principles are universal while perception is domain-specific drives a two-stage approach. By training control policies with privileged state in simulation and adapting only perception in the real world, we transform the complex sim-to-real problem into a structured perception alignment task.

Our experiments demonstrate the effectiveness of this decoupling. With only 10-20 real-world demonstrations, our method achieves performance comparable to end-to-end approaches using 4$\times$-8$\times$ more data. More importantly, the learned policies exhibit strong spatial generalization, maintaining a 35\% success rate in workspaces four times larger than the training region—something end-to-end methods struggle with. The gradual performance decay and contrast with baseline methods confirm that privileged state training enables robust, distance-invariant control strategies.

Beyond data efficiency, our approach enables modularity and stability by freezing the control policy after simulation training. This allows rapid deployment to new environments without retraining the entire system, addressing a key limitation of current sim-to-real methods. The structured perception learning problem also makes debugging and improvement easier compared to opaque end-to-end systems. Future work can extend this framework to more complex tasks, such as mobile manipulation and multi-arm coordination, and explore techniques for reducing real-world data requirements even further.

\input{bib}
\end{document}

%% file: intro.tex
\section{INTRODUCTION}

Simulation environments provide a safe, scalable, and cost-effective platform for robot learning~\cite{6}. We can run thousands of robots in parallel, automatically reset environments, and access perfect state information in simulation, making large-scale interactive learning feasible. However, reliably deploying simulation-trained policies to the real world remains a central challenge in robot learning~\cite{1,2,3}. This challenge not only concerns technical feasibility but fundamentally determines whether robot learning methods can transition to practical applications. The difficulty of sim-to-real transfer arises from the entanglement of perception and control.

In end-to-end learning paradigms, policies must simultaneously handle visual domain shift and dynamics gap. The former arises from discrepancies in texture, lighting, and appearance between simulated and real images, while the latter stems from differences in physical parameters such as friction, mass, and contact forces between simulation and reality~\cite{1,2,3,21,22}. This dual challenge creates a compounding effect: perceptual uncertainties interact with control uncertainties, making the sim-to-real gap grow multiplicatively rather than additively. At deployment, this entanglement makes policies brittle to real-world variations, often requiring extensive fine-tuning to achieve usable performance. 

Existing sim-to-real methods predominantly adopt end-to-end learning paradigms, attempting to train unified networks that directly map from pixels to actions~\cite{13,14,15,16,17,18,23,24,26}. While techniques like domain randomization have improved transfer performance~\cite{1,2,3}, they do not address the fundamental issue of perception-control entanglement. When perception and control are learned jointly, the network must master both visual feature extraction and control strategy generation simultaneously. This coupling means that even simple control behaviors require substantial data to learn under varying visual conditions~\cite{21,22}, as the network cannot separate task-relevant control patterns from domain-specific visual features. Moreover, errors in either perception or control can propagate through the network, making failures difficult to diagnose and correct. 

While privileged state learning has shown success in locomotion through teacher-student distillation~\cite{7,19,25}, manipulation presents distinct challenges. Unlike locomotion where proprioception often suffices, manipulation critically depends on precise visual perception for object localization and grasping~\cite{8,9,10,11,12}. Rather than distilling both perception and control in simulation, we advocate learning perception only in the target domain where visual observations are authentic. 
\begin{figure*}[!t]
\centering
\includegraphics[width=0.8\textwidth]{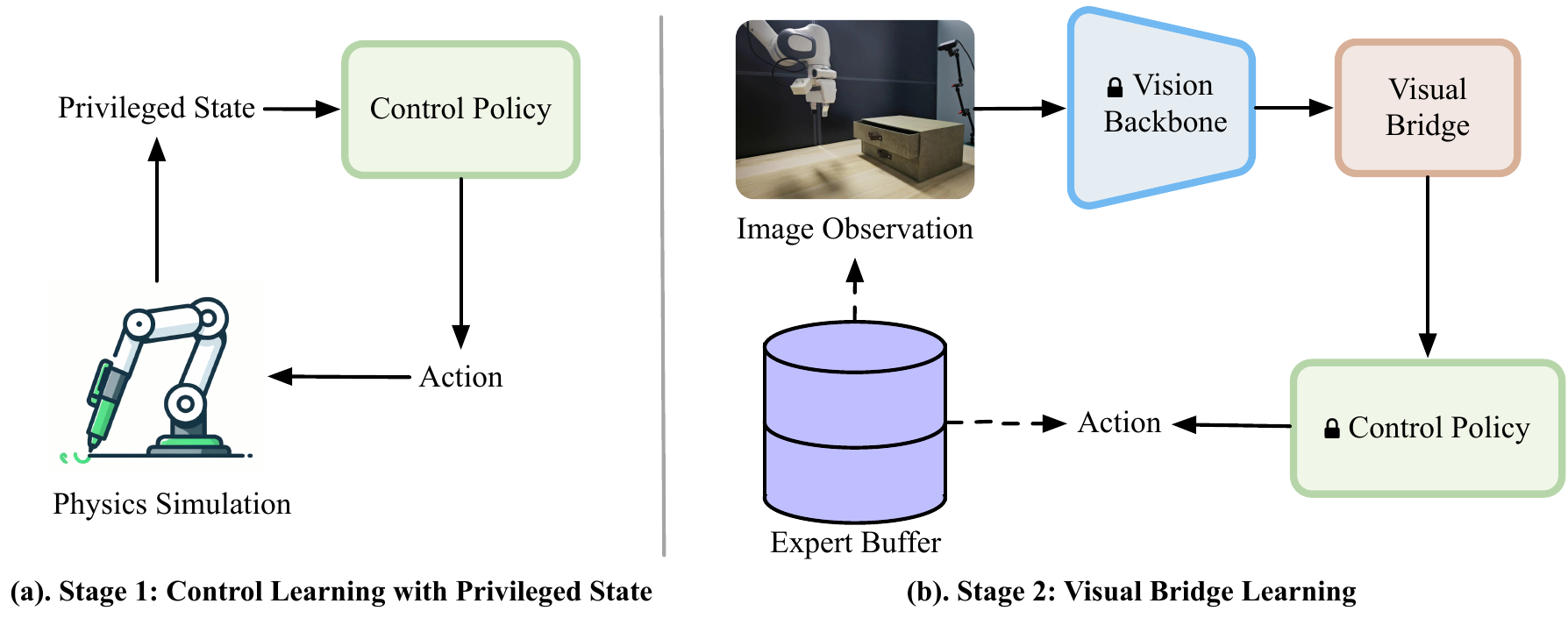}
\caption{Overview of our Best of Sim and Real (BSR) framework. (a) Stage 1: Control learning with privileged state in physics simulation, where the policy learns robust action patterns through systematic domain randomization. (b) Stage 2: Visual bridge learning in the real world, where a lightweight network maps image observations to the frozen control policy's input space using expert demonstrations stored in a replay buffer.}
\label{fig:overview}
\end{figure*}
Our core insight can be summarized as: \textbf{control is consistent, perception is specific}. Control strategies and action patterns are governed by consistent physical principles that remain invariant across environments, and they can therefore be effectively acquired in simulation through systematic randomization~\cite{1,2,3,5,6}. In contrast, perception is intrinsically tied to domain characteristics, since lighting conditions, textures, and visual appearances vary substantially across deployment scenarios and cannot be fully captured in simulation~\cite{8,10,11,12}. This perspective motivates a rethinking of sim-to-real transfer: instead of pursuing end-to-end policies that attempt to handle all variations simultaneously, we advocate decomposing the problem and training each component in the setting where it can be learned most effectively~\cite{20}.


Based on this insight, we propose \textbf{Best of Sim and Real (BSR): learn control where physics is accessible, adapt perception where visual observations are realistic.} In the first stage, we train control policies in simulation using privileged state—perfect object poses and spatial relationships—allowing the policy to focus purely on learning robust control strategies through systematic randomization~\cite{1,2,4,5,6}. In the second stage, we freeze the control policy and train only a lightweight visual bridge that maps real observations to the policy’s expected input space~\cite{10,11,12}. This decomposition transforms the complex sim-to-real problem into two well-defined subproblems: learning universal control patterns in simulation and solving a structured perception alignment task in the real world. 

We validate our approach on tabletop manipulation tasks, demonstrating superior data efficiency and out-of-distribution generalization compared to end-to-end baselines~\cite{13,14,15,16,17,18}. The learned policies generalize successfully to object positions and scales beyond the training distribution, highlighting that decoupling perception from control substantially improves sim-to-real transfer. More importantly, the learned policies exhibit strong generalization capabilities, handling object positions and scales outside the training distribution. Our main contributions are threefold: (1) proposing a new sim-to-real paradigm for manipulation that decouples perception and control, fundamentally reducing real-world data requirements; (2) designing a two-stage training framework based on privileged state that maximizes the advantages of both simulation and real environments; (3) systematic experimental validation of our method's advantages in data efficiency and generalization capability, with detailed ablation studies.

%% file: method.tex
\section{METHOD}

\subsection{Overview}

Since control is governed by invariant physical laws while perception is tied to environment-specific factors, we reformulate sim-to-real transfer as two separable subproblems, solvable in their respective domains.

We structure our framework into two complementary stages, as illustrated in Figure~\ref{fig:overview}. In the first stage (Section~\ref{sec:stage1}), we exploit the perfect state observability in simulation to train a control policy that masters the geometric and dynamic patterns of manipulation tasks. This policy learns from privileged state information---precise object poses, contact states, and relative transformations---allowing it to focus purely on the control problem without the confounding effects of perceptual noise. Once trained, this policy is frozen and serves as a fixed control backbone. In the second stage (Section~\ref{sec:stage2}), we address perception exclusively in the real world where visual observations are genuine. Rather than attempting to learn visual features in simulation where appearance realism is limited, we train a lightweight visual bridge network using a small number of real-world demonstrations to map camera observations to the representation space expected by the frozen control policy. This decoupling transforms the complex sim-to-real problem into a structured perception task with a clear learning target.

\begin{figure*}[t]
\centering
\includegraphics[width=0.85\textwidth]{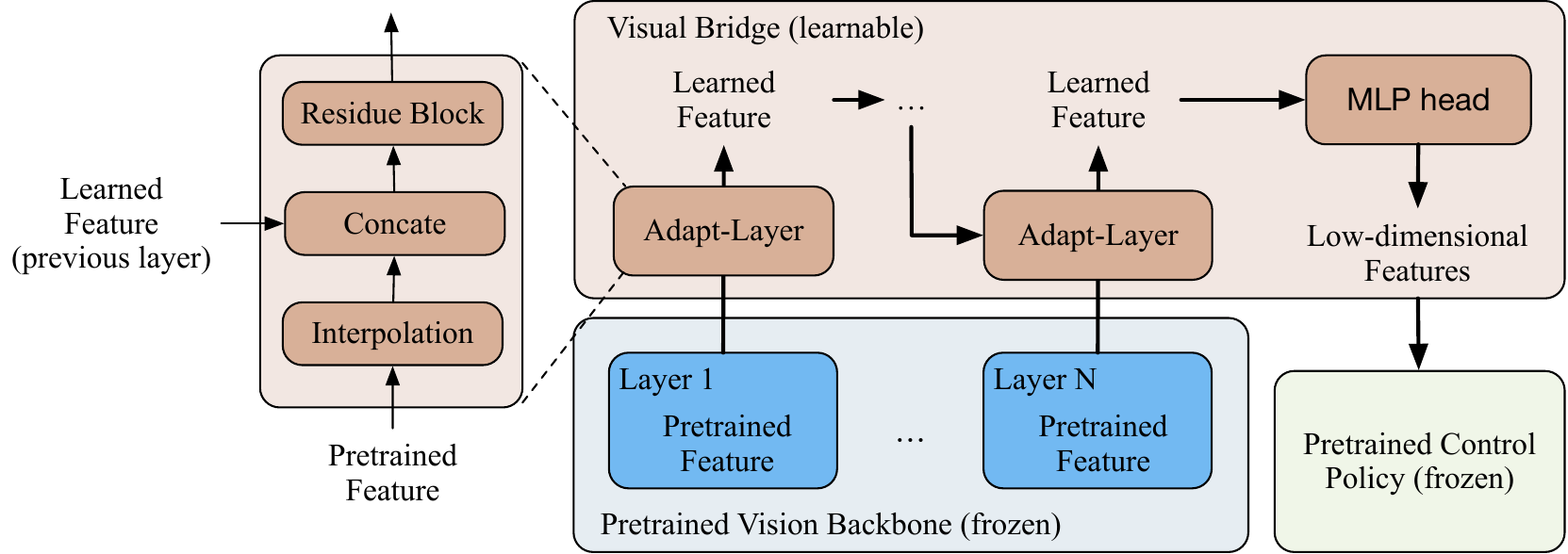}
\caption{Architecture of the visual bridge network. Multi-layer pretrained features from a frozen vision backbone are progressively refined through adaptive layers and residual blocks, then combined into low-dimensional features via an MLP head before being passed to the frozen control policy.}
\label{fig:visual_bridge}
\end{figure*}

\subsection{Stage 1: Control Learning with Privileged State}
\label{sec:stage1}

By providing the policy with ground-truth state information during simulation training, we enable it to focus exclusively on learning robust control strategies without perceptual uncertainties. Our privileged state representation consists of the robot's proprioceptive information including joint angles and velocities, the end-effector's 6-DoF pose, and task-relevant geometric state such as object poses expressed as relative transformations---the transformation from end-effector to object, the distance to grasp points, and approach alignments. This relative representation ensures that learned control patterns generalize across different spatial configurations. The policy network $\pi_\theta(a|s)$ then outputs action distributions for robot control.

To ensure robust transfer to the real world, we employ systematic domain randomization by incrementally intensify the extent of variation. Early in training, the policy experiences minor variations in object placement and physical parameters, allowing it to first acquire the basic task structure. As training progresses, we gradually expand both the dimensionality and magnitude of randomization. Geometric randomization encompasses variations in object positions, orientations, and scales, as well as diverse initial robot configurations. Physical randomization includes noise injection in observation and action spaces, variations in mass and friction coefficients, and explicit modeling of control delays. This progressive strategy ensures the policy develops robustness while maintaining stable learning dynamics. The simulation training leverages parallel simulators~\cite{6} to generate diverse experience efficiently, training the policy using PPO~\cite{5} with task-specific reward functions.

\subsection{Stage 2: Visual Bridge Learning}
\label{sec:stage2}

At the deployment phase, we face a structured learning problem: aligning real-world visual observations with the state representation required by the fixed control policy. This formulation fundamentally changes the nature of sim-to-real transfer from an unbounded reinforcement learning problem to a supervised learning task with a clear target. The frozen control policy acts as a strong prior, embodying the geometric and dynamic knowledge necessary for manipulation, while the perception problem is reduced to learning an appropriate visual bridge between observations and the control-relevant state space.

Our visual bridge leverages pretrained visual representations to provide strong perceptual priors, crucial for data-efficient learning. We employ a frozen vision backbone, specifically a Vision Transformer pretrained with self-supervised objectives like DINOv2~\cite{8}, which provides rich visual features without task-specific training. As illustrated in Figure~\ref{fig:visual_bridge}, we extract intermediate representations from multiple layers of the network. These multi-scale features capture both high-level semantic information and fine-grained spatial details, which are essential for robot manipulation. 


The bridge network processes these multi-layer features through adaptive layers that project them to a common dimensionality, followed by spatial alignment through bilinear interpolation to handle varying feature map resolutions~\cite{56}. We progressively fuse these representations through shallow residual blocks, where residual connections ensure stable gradient flow~\cite{23}. Then the fused visual representation is combined with proprioceptive signals through an MLP head, producing low-dimensional features suitable for the control policy.

Training the visual bridge requires only a small set of real-world demonstrations, typically 10-20 trajectories per task. During these demonstrations, we record synchronized camera images from both third-person and wrist-mounted viewpoints, proprioceptive readings, and expert actions. The bridge is trained to minimize the L2 distance between expert actions and those produced by the frozen control policy:
\begin{equation}
\mathcal{L} = \mathbb{E}_{(o_t, a_t^*) \sim \mathcal{D}_\text{demo}} \left[ \|a_t^* - \pi_\theta(f_\phi(o_t))\|^2 \right]
\end{equation}
where $f_\phi$ denotes the visual bridge, $o_t$ represents the observation, and $a_t^*$ is the expert action. This end-to-end supervision ensures the bridge learns to extract precisely the state information needed for successful task execution. The visual bridge contains only a small number of learnable parameters, focusing the learning problem and preventing overfitting on limited data.

\subsection{Best of Sim and Real Pipeline}
\label{sec:dst}

Our complete BSR pipeline coordinates the two-stage learning process to maximize both simulation efficiency and real-world performance. During the simulation phase, we establish task structure and reward design, then progressively introduce randomization following the curriculum described in Section~\ref{sec:stage1}. The policy typically converges within 10-20 million environment steps through parallel simulation. Once the policy demonstrates robust performance across the full randomization distribution, we freeze all parameters and export the trained model for deployment.

The real-world deployment phase begins with collecting expert demonstrations. Our system employs a dual-camera setup with calibrated extrinsics, maintaining consistent geometry between simulation and real deployment to minimize spatial domain shift. During visual bridge training, we apply data augmentation including random crops and color jittering to improve generalization. The complete pipeline enables practical deployment of new manipulation tasks with minimal real-world data requirements, as we demonstrate in Section~\ref{sec:experiments}.

%% file: experiment.tex
\section{Experiments}
\label{sec:experiments}
\begin{figure*}[t]
\centering
\includegraphics[width=\textwidth]{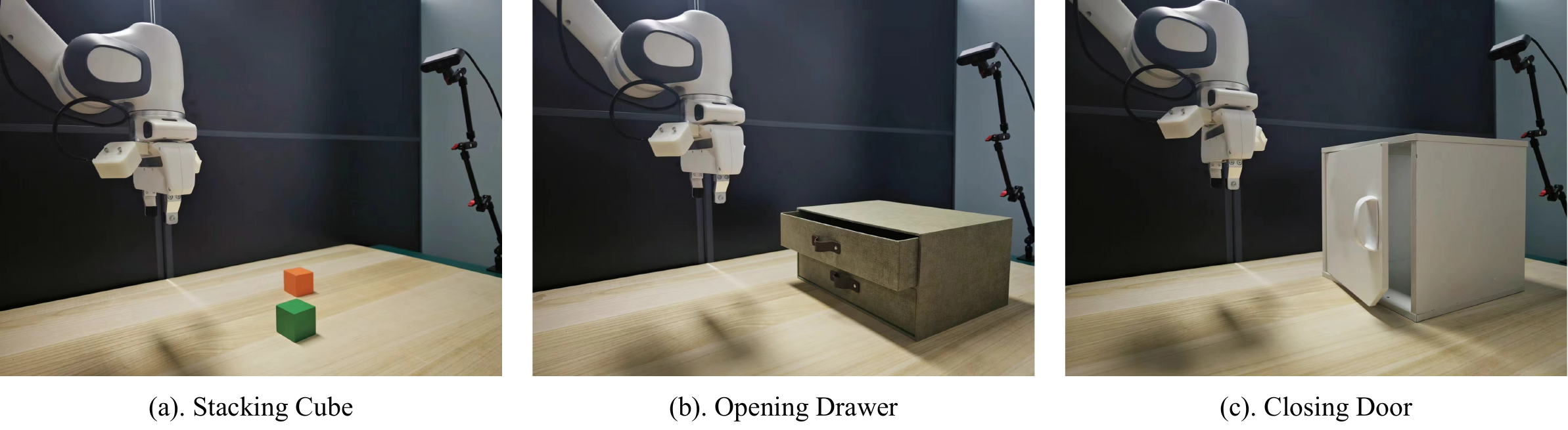}
\caption{Manipulation tasks used for evaluation. (a) \textbf{Stacking Cube}: Pick and place a cube onto a target platform, requiring precise grasp and placement within a 20$\times$20cm workspace. (b) \textbf{Opening Drawer}: Localize the handle, grasp, and pull to open the drawer by at least 15cm, demanding accurate visual servoing and force control. (c) \textbf{Closing Door}: Push a hinged door from 90° open to fully closed while maintaining continuous contact, testing the policy's ability to handle constrained motion and contact dynamics.}
\label{fig:tasks}
\end{figure*}

We design our experiments to validate three core claims: (1) decoupling perception and control fundamentally improves sim-to-real transfer efficiency, (2) learning control with privileged state in simulation provides superior spatial generalization compared to end-to-end learning, and (3) our visual bridge design enables effective transfer with minimal real-world data. We evaluate on three manipulation tasks that require precise visual-motor coordination: \textbf{Stacking Cube} (pick and stack objects with varying sizes), \textbf{Opening Drawer} (localize handle and execute constrained trajectory), and \textbf{Closing Door} (swing door requiring continuous contact control).

\begin{figure*}[t]
    \centering
    \includegraphics[width=\textwidth]{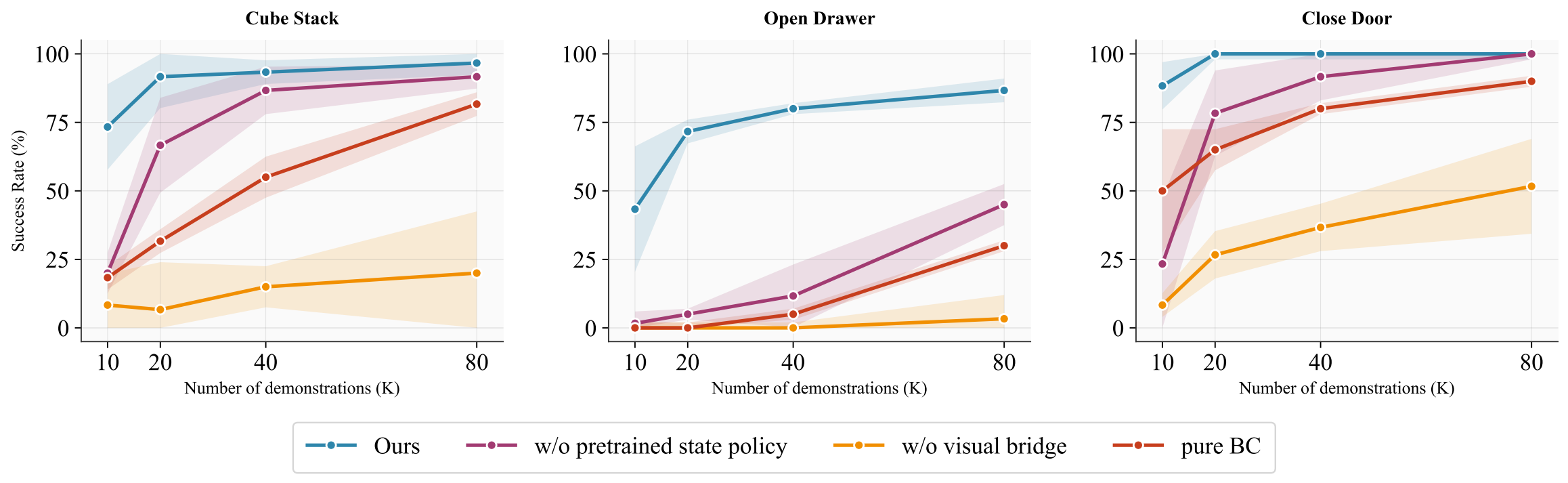}
    \caption{Success rate as a function of real-world demonstrations. Our method achieves strong performance with just 10–20 demonstrations, while baselines require substantially more data or fail to reach comparable performance.}
    \label{fig:sample_efficiency}
\end{figure*}

\subsection{Experimental Setup}

\textbf{Tasks and Metrics.} We evaluate our approach on three manipulation tasks illustrated in Figure~\ref{fig:tasks}. For Stacking Cube (Fig.~\ref{fig:tasks}a), robots must pick up a cube and stack it on a target platform within a 20cm$\times$20cm workspace. Opening Drawer (Fig.~\ref{fig:tasks}b) requires localizing and grasping a handle, then pulling to open a drawer by at least 15cm. Closing Door (Fig.~\ref{fig:tasks}c) involves pushing a hinged door from 90° open to fully closed while maintaining continuous contact. We report success rate as the primary metric and additionally use a graded completion score 0-4 to capture partial task progress for detailed analysis. For Stacking Cube, the completion score indicates: 0 (failure), 1 (approach), 2 (grasp), 3 (move to target), and 4 (successful stack).

\textbf{Baselines.} We compare against three ablations that isolate key design choices: (1) \textit{w/o pretrained state policy}: End-to-end learning from pixels to actions without leveraging simulation-trained control; (2) \textit{w/o visual bridge}: Direct state regression from images instead of representation-level bridging; (3) \textit{pure BC}: Behavior cloning without pretrained vision encoder, learning visual features from scratch. All methods use identical network capacity, training budgets, and data augmentation for fair comparison.

\subsection{Sample Efficiency in Real World}

Figure~\ref{fig:sample_efficiency} presents our main result on sample efficiency. With only 10 real-world demonstrations per task, our method achieves 73.3\% success on Stacking Cube, 43.3\% on Opening Drawer, and 88.3\% on Closing Door. In contrast, the strongest baseline (w/o pretrained state policy) only reaches 20.0\%, 1.7\%, and 50.0\% respectively---a gap of 30-50 percentage points. This significant difference validates our core thesis: by learning control in simulation and only adapting perception in the real world, we fundamentally reduce the complexity of real-world learning.

The performance gap persists but narrows as more demonstrations become available. At $K=80$, end-to-end learning approaches our performance on some tasks, but requires 4$\times$-8$\times$ more data to reach the same success rate we achieve at $K=10$-$20$. Notably, removing the visual bridge (green curve) severely degrades performance across all data regimes, confirming that representation-level bridging is crucial for connecting visual observations to the frozen control policy. Pure behavior cloning without pretrained encoders (red curve) shows the poorest performance, emphasizing the importance of visual priors for few-shot learning.

\subsection{Spatial Generalization Beyond Training}

\begin{figure*}[t]
\centering
\includegraphics[width=\textwidth]{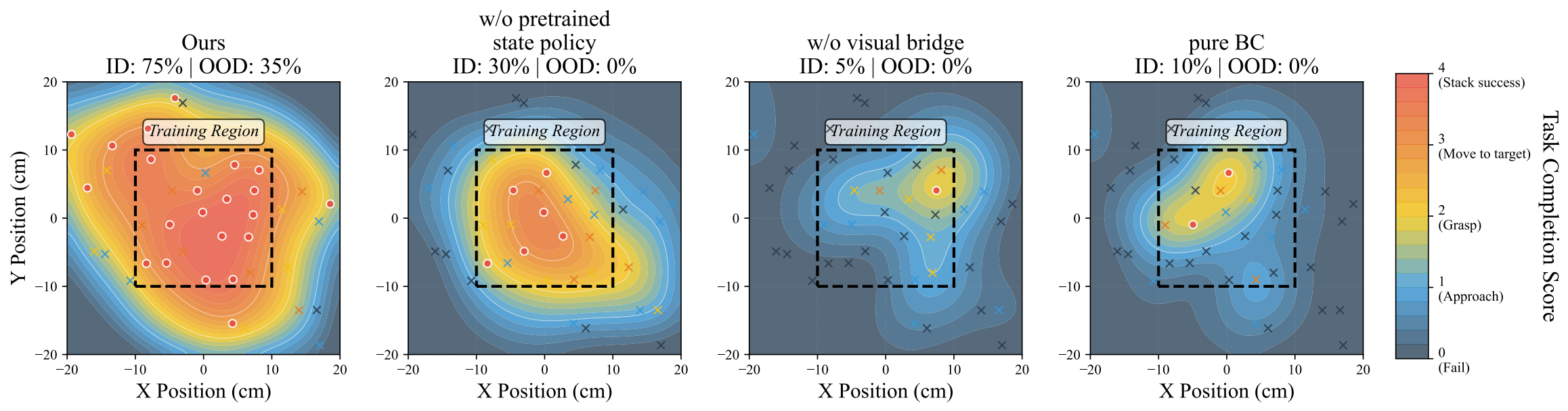}
\caption{Spatial visualization of task completion scores for Stacking Cube as the workspace expands from 20$\times$20cm training region (black dashed box) to 40$\times$40cm evaluation area. Heatmaps show interpolated completion scores (0-4 scale), with circles indicating successful trials and crosses marking failures. Our method maintains high performance (ID: 75\%, OOD: 35\%) across the extended workspace, while baselines show rapid degradation beyond the training boundary.}
\label{fig:spatial_heatmap}
\end{figure*}

A key advantage of learning universal control principles in simulation is enhanced spatial generalization. To rigorously assess this capability, we conduct extensive experiments on the Stacking Cube task, expanding the workspace from the original 20cm$\times$20cm training region to a 40cm$\times$40cm evaluation area, representing a fourfold increase in area. We define in-distribution (ID) performance as success within the original training region, and out-of-distribution (OOD) performance as success in the outer regions beyond the training boundary.

Figure~\ref{fig:spatial_heatmap} provides detailed spatial visualizations of task completion scores across the extended workspace. Our method achieves 75\% ID success rate and maintains 35\% OOD success rate, demonstrating meaningful generalization beyond the training distribution. The heatmap reveals that our approach maintains high completion scores (3-4, indicating successful grasping and stacking) throughout most of the workspace, with gradual degradation toward extreme corners. This spatial consistency confirms that the privileged state policy learns genuine geometric and dynamic patterns rather than memorizing specific configurations.

In stark contrast, baseline methods show catastrophic failure beyond the training boundary. The end-to-end approach without pretrained state policy drops from 30\% ID to 0\% OOD success, with completion scores rapidly declining to 0-1 (failure to approach or grasp) outside the training region. Similarly, methods without visual bridge or pretrained encoders achieve only 5\% and 10\% ID success respectively, with complete failure in OOD regions. These results highlight a fundamental limitation of end-to-end learning: when perception and control are entangled, the policy overfits to the specific visual patterns seen during training, preventing generalization to new spatial configurations.

\begin{figure}[t]
\centering
\includegraphics[width=\columnwidth]{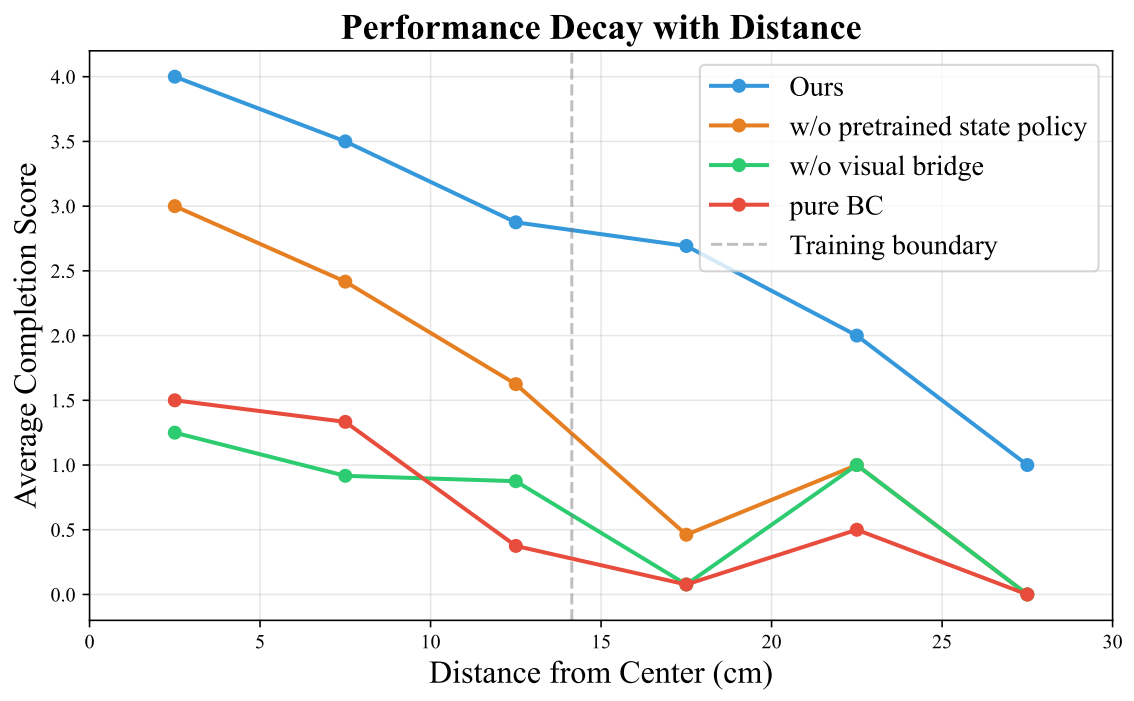}
\caption{Performance decay as a function of distance from the training region center. Our method shows graceful degradation with distance, maintaining an average completion score above 2.0 even at 22.5cm (beyond training boundary at 14.1cm), while baselines exhibit sharp performance cliffs.}
\label{fig:performance_decay}
\end{figure}

Figure~\ref{fig:performance_decay} quantifies this generalization pattern by plotting average completion scores as a function of distance from the workspace center. The training boundary at 14.1cm (diagonal distance from center to corner of 20cm$\times$20cm region) is marked with a dashed line. Our method exhibits graceful degradation, maintaining scores above 2.0 (successful grasping) even at 22.5cm from center. The gradual slope indicates that our decoupled approach learns distance-invariant control strategies through privileged state training.

Baselines show markedly different behavior with sharp performance cliffs near the training boundary. The w/o pretrained state policy baseline drops from 3.0 to below 0.5 beyond 15cm, indicating complete task failure. Pure BC and w/o visual bridge methods perform poorly even within the training region and show near-zero performance outside. These sharp transitions suggest that end-to-end methods learn spurious correlations between absolute positions and actions rather than relative geometric relationships that generalize.

The superior spatial generalization of our method directly stems from the decoupled training paradigm. By learning control with privileged state in simulation, the policy focuses on relative transformations---end-effector to object distances, approach angles, grasp configurations---that remain valid regardless of absolute position. The systematic domain randomization during simulation training further ensures robustness to spatial variations. Meanwhile, the visual bridge, trained with limited real demonstrations, only needs to extract these relative features from images rather than learning position-dependent control strategies.

\subsection{Ablation Studies}

\begin{table}[t]
\centering
\caption{Ablation Study on Visual Bridge Components ($K=20$)}
\label{tab:ablation}
\begin{tabular}{lccc|c}
\hline
Component & Stack & Drawer & Door & Avg \\
\hline
Full model & 91.7 & 71.7 & 100.0 & 87.8 \\
Single layer features & 78.3 & 55.0 & 91.7 & 75.0 \\
w/o residual connections & 83.3 & 60.0 & 95.0 & 79.4 \\
w/o proprioception & 71.7 & 48.3 & 88.3 & 69.4 \\
Direct FC mapping & 65.0 & 41.7 & 85.0 & 63.9 \\
\hline
\end{tabular}
\end{table}

As shown in Table~\ref{tab:ablation}, we conduct a detailed analysis of the contribution of each visual bridge component to the model’s performance. First, we observe a significant reduction in average success (12.8\%) when we use only the final-layer features for the task, instead of multi-scale feature extraction. This suggests that intermediate layers play a critical role in capturing essential spatial details, which are vital for accurate manipulation across the various tasks. The reduction in success highlights the necessity of extracting spatial information at different levels of abstraction to handle complex manipulations effectively.

Next, we analyze the effect of removing residual connections, which results in an 8.4\% decrease in performance. This emphasizes the role of residual connections in maintaining optimization stability, particularly with limited data. The most substantial drop (23.9\%) occurs when replacing the progressive fusion architecture with direct fully-connected (FC) mapping. This outcome further validates our design choice to preserve spatial structure through gradual aggregation, as direct FC mapping fails to capture essential spatial patterns, leading to a significant degradation in task performance.

\subsection{Discussion}

Our experiments demonstrate that decoupling perception and control fundamentally changes the sample complexity of sim-to-real transfer. The 4$\times$-8$\times$ improvement in data efficiency stems from transforming an unbounded RL problem into a structured supervised learning task. More importantly, the strong spatial generalization---maintaining 35\% success rate in regions four times larger than training---confirms that simulation-trained policies with privileged state capture genuine task structure rather than memorizing configurations. The gradual performance decay with distance, as opposed to sharp cliffs in baselines, validates that our approach learns robust, distance-invariant control strategies that transfer across spatial scales. These results suggest that the key to practical sim-to-real transfer lies not in perfecting simulation fidelity, but in identifying what to learn where: control in simulation where physics is accessible, perception in the real world where appearance is authentic.

%% file: related.tex
\section{Related Work}

\subsection{Sim-to-Real Transfer in Robot Manipulation}

Sim-to-real transfer remains a fundamental challenge in robot learning, with multiple approaches proposed to bridge the gap between simulation and real physical world. Domain randomization (DR) methods \cite{1, 2, 3} train policies on diverse simulated environments to achieve effective transfer. Tobin et al. \cite{1} pioneered this approach by randomizing visual parameters, while Peng et al. \cite{2} extended it to dynamics randomization. Furthermore, Chebotar et al. \cite{3} proposed closing the sim-to-real loop by adapting randomization parameters based on real-world experience. Recent extensions \cite{27, 28} have shown that curriculum learning-like design in randomization improves transfer, with \cite{29} demonstrating impressive dexterous manipulation through systematic randomization and \cite{30} solving complex tasks like Rubik's cube.

Different from DR methods, domain adaptation techniques attempt to align simulation and real distributions through learnable strategies. Asymmetric architectures \cite{4} separate observation and state processing for improved transfer. Recent work explores adversarial domain adaptation \cite{31}, style transfer \cite{32}, and progressive networks \cite{33} that prevent catastrophic forgetting. Meta-learning approaches \cite{34, 35} enable rapid adaptation, while system identification methods \cite{36, 37} estimate real-world parameters to improve simulation fidelity.

Hybrid approaches combine multiple strategies for robustness. \cite{38} demonstrated sim-to-sim transfer before real deployment, while \cite{39} showed that self-supervised adaptation improves visual robotic manipulation. Despite these advances, existing methods treat perception and control as entangled problems, requiring extensive real-world data. Our work fundamentally differs by completely decoupling these components across domains, learning control with privileged state in simulation while adapting only perception in the real world with minimal demonstrations.

\subsection{Learning with Privileged Information and Visual Representations}

Privileged information has emerged as a powerful paradigm in robot learning. Kumar et al. \cite{7} demonstrated rapid motor adaptation for legged robots through teacher-student distillation from privileged to deployable policies. Recent work \cite{19} extended this with privileged sensing scaffolds that guide reinforcement learning. Zeng et al. \cite{25} applied similar concepts to visual legged locomotion, learning world models from privileged state. In manipulation, \cite{40} leveraged privileged depth information for improved learning, while \cite{41} used privileged force feedback for contact-rich tasks. However, these approaches typically distill all components in simulation, missing the opportunity to learn perception where visual observations are authentic.

The quality of visual representations critically impacts manipulation performance. Self-supervised pretraining has produced certain powerful foundation models. R3M \cite{10} learns from human interaction videos, VC-1 \cite{11} provides universal visual representations for embodied AI, while SUGAR \cite{12} pre-trains 3D representations specifically for robotics. DINOv2 \cite{8} offers robust features through self-supervised learning, building on the Vision Transformer architecture \cite{9}. Recent work includes MVP \cite{42} using masked autoencoding, CLIP \cite{43} for vision-language alignment, and specialized representations for robotic tasks \cite{44, 45}.

Our approach uniquely leverages these pretrained representations as a bridge between real observations and simulation-trained control policies. Unlike prior work that uses pretrained models within end-to-end frameworks \cite{13, 14, 15, 16}, we employ them specifically for perception alignment, transforming the complex sim-to-real problem into a structured supervised learning task requiring only 10-20 real demonstrations.

\subsection{End-to-End Visuomotor Learning and Modular Approaches}

End-to-end visuomotor learning has achieved impressive results through scale and architectural innovations. Previous works use large-scale datasets to enable training generalizable policies: BridgeData V2 \cite{14} provides diverse manipulation demonstrations, DROID \cite{15} offers in-the-wild robot data, and RoboNet \cite{46} enables multi-robot learning. Vision-language-action models have emerged as a powerful paradigm, with RT-2 \cite{16} transferring web knowledge to robotic control, RT-1 \cite{47} demonstrating real-world scaling, and PaLM-E \cite{24} providing embodied multimodal capabilities. OpenVLA \cite{26} advances general-purpose vision-language-action models, while cross-embodiment initiatives like Open X-Embodiment \cite{17} and Octo \cite{18} pursue generalist policies through massive data aggregation.

Architectural innovations have improved end-to-end learning efficiency. Transformer-based policies \cite{48, 49} model long-horizon dependencies, while Perceiver-based architectures \cite{50} handle multimodal inputs effectively. Data augmentation techniques partially address sample efficiency: RAD \cite{21} showed simple augmentations improve RL, while DrQ-v2 \cite{22} achieved state-of-the-art continuous control through aggressive augmentation. However, the fundamental coupling of perception and control limits their efficiency compared to our decoupled approach.

Modular approaches have explored various decomposition strategies. Rizzardo et al. \cite{20} proposed latent prediction for non-prehensile manipulation, while \cite{51} separated visual encoding from policy learning. Hierarchical decomposition \cite{52, 53} addresses long-horizon tasks, and skill-based methods \cite{54, 55} enable compositional learning. Our work advances modularity by completely decoupling perception and control across domains, achieving 4-8× better data efficiency than end-to-end baselines while maintaining strong spatial generalization—a key advantage demonstrated by our policies successfully handling workspaces four times larger than the training region.